\documentclass[a4paper,fleqn]{cas-sc}

\usepackage[numbers]{natbib}
\usepackage{graphicx}%
\usepackage{multirow}%
\usepackage{amsmath,amssymb,amsfonts}%
\usepackage{amsthm}%
\usepackage{mathrsfs}%
\usepackage{xcolor}%
\usepackage{booktabs}%
\usepackage{algorithm}%
\usepackage{algorithmicx}
\usepackage{algpseudocode}%

\def\tsc#1{\csdef{#1}{\textsc{\lowercase{#1}}\xspace}}
\tsc{WGM}
\tsc{QE}

\begin{document}
\let\WriteBookmarks\relax
\def\floatpagepagefraction{1}
\def\textpagefraction{.001}

\shorttitle{Webly Supervised Multi-Label Recognition}    

\shortauthors{Zhihua Xu et al.}  

\title [mode = title]{Webly Supervised Multi-Label Recognition: Evaluation Benchmark and Dual-Branch Multi-Label Contrastive Learning}  

\author[1]{Zhihua Xu}[orcid=0000-0002-0731-4585]
\ead{xuzhh76@mail2.sysu.edu.cn}
\affiliation[2]{organization={Guangdong University of Technology},
            city={Guangzhou},
            postcode={510006}, 
            state={GuangDong},
            country={China}}
\credit{Methodology, Software, Validation, Data curation, Writing - Original draft preparation, Writing - Review \& Editing}

\author[2]{Zhijing Yang}
\ead{yzhj@gdut.edu.cn}

\credit{Resources, Funding acquisition, Project administration}



\author[1]{Yufeng Yang}
\cormark[1]
\ead{yangyf226@mail.sysu.edu.cn}
\affiliation[1]{organization={Sun Yat-Sen University},
            city={Guangzhou},
            postcode={510006}, 
            state={GuangDong},
            country={China}}
\credit{Data curation, Formal analysis, Writing - Review \& Editing}

\author[2]{Tianshui Chen}
\ead{tianshuichen@gmail.com}
\credit{Conceptualization, Methodology, Supervision, Project administration, Writing - Review \& Editing}

\cortext[1]{Corresponding author}

\begin{abstract}
Training deep learning models with freely available web images can reduce their dependence on costly manual annotations. Although webly supervised learning has been widely studied for single-label recognition, its multi-label counterpart remains underexplored, partly due to the lack of unified benchmarks and fair comparison protocols. To address this gap, we construct a benchmark for webly supervised multi-label recognition (WS-MLR), including Web-COCO and Web-Pascal, and re-implement representative baselines under a unified setting. The two datasets cover the same 80 and 20 categories as MS-COCO and Pascal VOC, respectively, and contain about 300 thousand images retrieved from the Internet using category-word combinations as search keywords. We further propose a Dual-Branch Multi-Label Contrastive Learning (DBMLCL) framework, which learns category-specific instance-level and category-level representations together with their similarities to identify and correct noisy labels. Extensive experiments on the benchmark demonstrate that DBMLCL achieves superior performance compared to representative baselines.
\end{abstract}



\begin{keywords}
Multi-Label Recognition\sep Webly Supervised\sep Contrastive Learning\sep Evaluation Benchmark
\end{keywords}

\maketitle
\section{Introduction}
\label{sec:intro}
Multi-label image recognition (MLR) is more fundamental and practical than its single-label counterpart since a real-world image inherently contains multiple objects belonging to diverse semantic categories. Despite achieving impressive progress, recent MLR methods \cite{wei2016hcp,chen2019learning,CAO2021104238} mainly depend on data-hungry deep learning paradigms and require large-scale manually annotated datasets. Constructing well-annotated large-scale MLR datasets, such as Microsoft COCO \cite{lin2014microsoft} and Pascal VOC \cite{everingham2010pascal}, is extremely expensive and difficult due to the complex label and image spaces.

Recently, more studies have resorted to MLR with partial labels \cite{huynh2020interactive,chen2022structured,chen2024heterogeneous} and few-shot MLR \cite{alfassy2019laso,chen2022knowledge} to alleviate this dilemma. These works still require various forms of human intervention and thus remain labor-intensive for data collection. To further reduce annotation costs, increasing efforts \cite{chen2015webly,li2020mopro,sun2021webly} are dedicated to webly supervised classification, which searches large-scale data from the Internet and uses them to train deep classification models \cite{he2016deep,simonyan2015very,chen2021cross,pu2023spatial}. Existing works mainly focus on single-label image classification, which assumes that each image contains objects belonging to a specific category, and only a few algorithms target its multi-label counterpart. The lack of evaluation benchmarks, including publicly available datasets and comparison algorithms, makes it difficult to evaluate the performance of different algorithms.

Therefore, we first construct a unified evaluation benchmark for webly supervised multi-label image recognition (WS-MLR), which unifies the datasets and baselines for fair evaluations. Concretely, because Microsoft COCO \cite{lin2014microsoft} and Pascal VOC \cite{everingham2010pascal} are the two most widely used datasets to evaluate the MLR task, we search hundreds of thousands of images from the Internet using combinations of category words from these two datasets as keywords. In this way, we obtain two webly retrieved datasets, i.e., Web-COCO and Web-Pascal. These two datasets contain about 300,000 images and cover the same 80 and 20 categories as Microsoft COCO and Pascal VOC, respectively. For comparisons, we further re-implement recent MLR, few-shot MLR, and partial-label MLR algorithms and adapt these algorithms to address the WS-MLR task. We present extensive and fair comparisons to evaluate the performance of each algorithm.

Label noise is the key challenge for the WS-MLR task, and it becomes more complex from single-label to multi-label scenarios. The space of label noise expands from $n$ to $2^n$ for an image with $n$ given labels because each label could be correct or incorrect. On the other hand, in contrast to single-label images, multi-label images commonly contain multiple semantic objects scattered over the whole image. It is also difficult to automatically locate the regions corresponding to each label, which further increases the difficulty. To address these issues, we propose a dual-branch multi-label contrastive learning (DBMLCL) framework, which learns category-specific instance-level and category-level feature representations, together with their similarities to help identify and correct noisy labels. Specifically, each image is fed into two backbone networks followed by a semantic-aware feature learning model to extract category-specific instance-level feature vectors and to update the category-level feature vectors. Then, for a specific category, we compute the similarity between instance-level and category-level feature vectors to estimate whether this category is present and thus help to correct noisy labels. During training, we introduce a category-specific instance-level and instance-category contrastive learning loss to learn compact category-level features and instance-category similarity. 

Our contributions can be summarized as follows. First, we construct an evaluation benchmark for WS-MLR, which consists of the newly constructed Web-COCO and Web-Pascal datasets, and re-implement several representative baselines for fair comparisons. This benchmark can help future studies better verify the effectiveness of newly proposed algorithms. Second, we propose a dual-branch multi-label contrastive learning framework that adapts contrastive learning and label correction to the WS-MLR setting by learning category-specific instance-level and category-level feature representations together with their similarities. This design helps correct noisy web labels at the category level and improves WS-MLR performance. Finally, we conduct experiments to compare the proposed framework with these baselines to demonstrate its effectiveness. The datasets, implementation codes of the proposed framework and baselines, and trained models are available at \url{https://github.com/zizizihua/WS-MLR}.

\section{Related Work}

\subsection{Multi-Label Image Recognition}
Multi-label image recognition (MLR) aims to recognize multiple semantic categories from a single image and has been widely applied to content-based image retrieval, recommendation systems, and visual understanding applications~\cite{zhao2015deep,shen2022deep,MA2022108216,DENG2024105319,WU2024105189,chen2026learning}. 
In the last decades, many methods have been proposed to improve category-specific representation learning and label dependency modeling~\cite{Wu2020AdaHGNN,Ye2020ADD-GCN,chen2021p-gcn,chen2022knowledge,ridnik2021asymmetric,Pu2022SRDL,chen2024dynamic,WANG2022104548}. 
For example, graph-based methods exploit label co-occurrence or semantic relations to propagate information among categories, while attention-based methods learn category-specific visual representations for multi-label prediction. 
Despite their effectiveness, most existing MLR methods rely on large-scale clean and complete multi-label annotations, which are expensive and time-consuming to collect, especially when the number of categories and images increases.

\subsection{Multi-Label Recognition with Incomplete or Noisy Labels}
To reduce annotation costs, several weakly supervised MLR settings have been studied. 
Multi-label few-shot learning (ML-FSL) aims to recognize novel categories with limited labeled samples~\cite{alfassy2019laso,chen2022knowledge,Simon2022meta}. 
For instance, Alfassy et al.~\cite{alfassy2019laso} synthesize multi-label samples through label-set operations, Chen et al.~\cite{chen2022knowledge} exploit prior knowledge to guide adaptive information propagation among categories, and Simon et al.~\cite{Simon2022meta} estimate the label count of a given sample through relational inference. 
Another related setting is MLR with partial labels, where only a subset of positive labels is observed and the remaining labels are unknown~\cite{durand2019learning,huynh2020interactive,chen2022structured}. 
Representative methods train classifiers with observed labels and then infer pseudo labels for unobserved categories, or mine intra-image and cross-image correlations to complement missing annotations. 

More closely related to our task is multi-label learning with noisy labels, where both false-positive and false-negative labels may exist. 
Kim et al.~\cite{Kim2022LargeLoss} regard unobserved labels as negative labels and cast weakly supervised MLR into a noisy multi-label learning problem, showing that neural networks may eventually memorize noisy labels and proposing large-loss rejection or correction to alleviate this issue. 
Beyond missing-label noise, recent studies explicitly consider more general noisy multi-label settings. 
For example, Zhao et al.~\cite{zhao2024ijcai-robust} formulate noisy multi-label learning with dirty label noise and decompose the corrupted label matrix into a true label component and a noise component, while Hou et al.~\cite{hou2025noisy} exploit label co-occurrence information to model and recover cleaner multi-label supervision. 
These studies suggest that robust MLR requires modeling label corruption at the instance-label or label-correlation level rather than treating all unobserved labels uniformly. 
However, they are usually designed for curated datasets with synthetic, partially observed, or post-hoc corrupted labels, while webly supervised MLR faces more complex noise patterns caused by image search, keyword mismatch, semantic ambiguity, and missing co-occurring objects. 
Therefore, directly applying existing incomplete-label or noisy-label MLR methods to webly collected multi-label data may lead to sub-optimal performance.

\subsection{Webly Supervised Learning}
Webly supervised learning aims to reduce annotation cost by collecting large-scale images from the Internet and using their associated tags or search keywords as weak supervision~\cite{chen2015webly,li2017webvision,mahajan2018exploring,sun2021webly}. 
Early studies assume that large-scale web data can compensate for label noise and directly train deep models with noisy web labels~\cite{li2017webvision,mahajan2018exploring}. 
However, noisy web labels can severely harm model training and lead to sub-optimal performance, especially when false-positive and false-negative labels are widely present. 
To improve robustness, recent works introduce contrastive learning or prototype-based learning to better exploit noisy web data~\cite{li2020mopro}. 

Most existing webly supervised methods focus on single-label classification or fine-grained recognition, where each image is usually associated with one dominant category. 
In contrast, webly supervised multi-label recognition (WS-MLR) is more challenging because each image may contain multiple semantic objects scattered across different regions, and the search keywords usually provide incomplete and noisy supervision. 
Existing webly supervised methods commonly perform contrastive learning on holistic image features, which can hardly distinguish different category-specific regions in multi-label images. 
Differently, our DBMLCL framework learns category-specific instance-level and prototype-level representations, which helps reduce semantic confusion and correct noisy web labels in the WS-MLR setting.

\subsection{Vision-Language Models for Multi-Label Recognition}
Recently, vision-language models such as CLIP~\cite{radford2021learning} have shown strong zero-shot recognition ability and have been adopted for weakly supervised, unsupervised, and open-vocabulary multi-label recognition. 
Since CLIP is originally trained with image-text contrastive learning and mainly produces global image-text matching scores, directly applying it to multi-label recognition may suffer from incomplete localization and inaccurate prediction of multiple co-existing categories. 
To address this issue, CDUL~\cite{Abdelfattah_2023_ICCV} proposes a CLIP-driven unsupervised learning framework for annotation-free multi-label image classification. 
It aggregates global and local image-text similarities generated by CLIP to initialize pseudo labels, and then trains a classification network while refining pseudo labels for unobserved labels. 
More recently, Classifier-guided CLIP Distillation (CCD)~\cite{Kim_2025_CVPR} further improves CLIP-based unsupervised multi-label classification by distilling pseudo labels from CLIP predictions with classifier-guided local views and prediction debiasing. 

These methods demonstrate that vision-language priors can provide useful category-level semantic knowledge without requiring manually annotated training labels. 
However, CLIP-based unsupervised MLR mainly focuses on generating and refining pseudo labels from vision-language similarity, while WS-MLR provides noisy web keyword labels that contain both missing positive labels and incorrect positive labels. 
Therefore, directly transferring CLIP-based pseudo-label distillation to WS-MLR may be insufficient, and explicit modeling of noisy web labels remains necessary.

\section{Evaluation Benchmark}
In this section, we present a WS-MLR evaluation benchmark, which consists of the newly constructed datasets and re-implemented baselines. 

\subsection{Datasets}

\noindent\textbf{Data Collection.}
We first determine the multi-label categories before collecting the WS-MLR datasets. Microsoft COCO \cite{lin2014microsoft} and Pascal VOC \cite{everingham2010pascal} are the two most widely used datasets to evaluate traditional MLR algorithms. These two datasets cover 80 and 20 categories, respectively, in which the 20 categories of the VOC dataset are a subset of the 80 categories of the COCO dataset. Thus, we use the 80 COCO categories to build the Web-COCO and Web-Pascal datasets. Then, we randomly select one or more categories as keywords (e.g., ``person'' or ``person, car, bus'') to search for the corresponding images from image search engines, including Google, Baidu, and Bing Image Search. In this way, we obtain more than 500,000 images. We then remove the broken and duplicated images and use the remaining 290,000 images to build the Web-COCO dataset. This filtering process is performed automatically by scripts. Specifically, we first check whether each downloaded file can be correctly parsed as a valid image file and discard files with invalid image formats. We then compute a hash value for each valid image file and remove duplicated images with identical hashes. We further select the subset that contains at least one of the 20 VOC categories to build the Web-Pascal dataset. 

\begin{figure}
\centering
\includegraphics[width=0.8\textwidth]{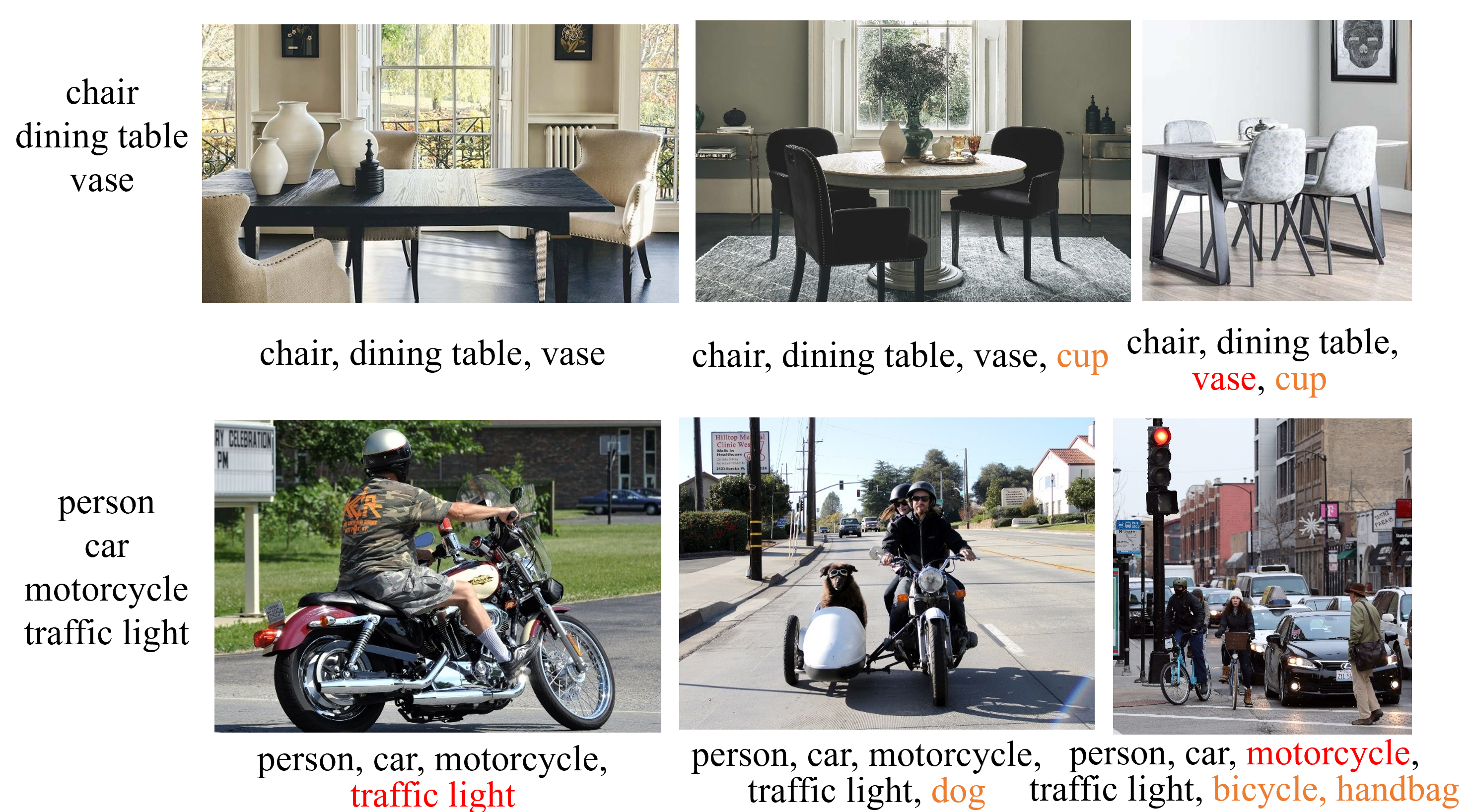}
\caption{Some examples with false-positive and false-negative labels. The keywords are presented on the left and the manually-annotated labels are presented under each image, with false-positive labels in red and false-negative labels in orange.}
\label{fig:samples}
\end{figure}

\begin{figure}
\centering
\includegraphics[width=0.99\textwidth]{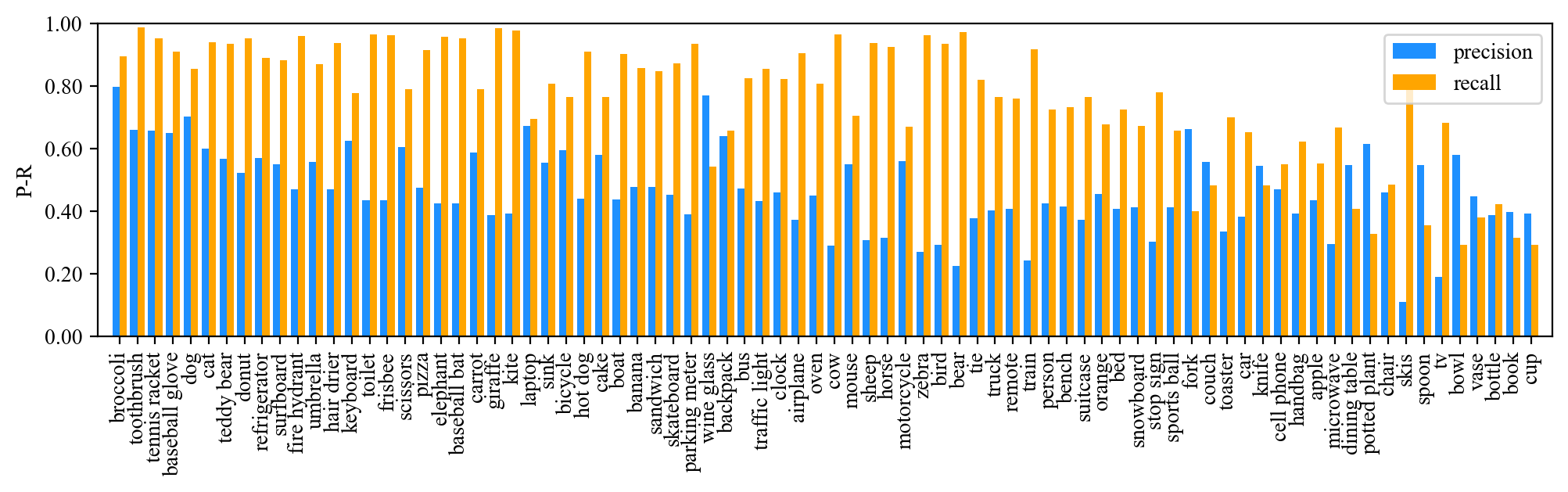}
\caption{Precision and recall of pseudo labels in the Web-COCO dataset.}
\label{fig:pr}
\end{figure}

\begin{figure}
\centering
\includegraphics[width=0.99\textwidth]{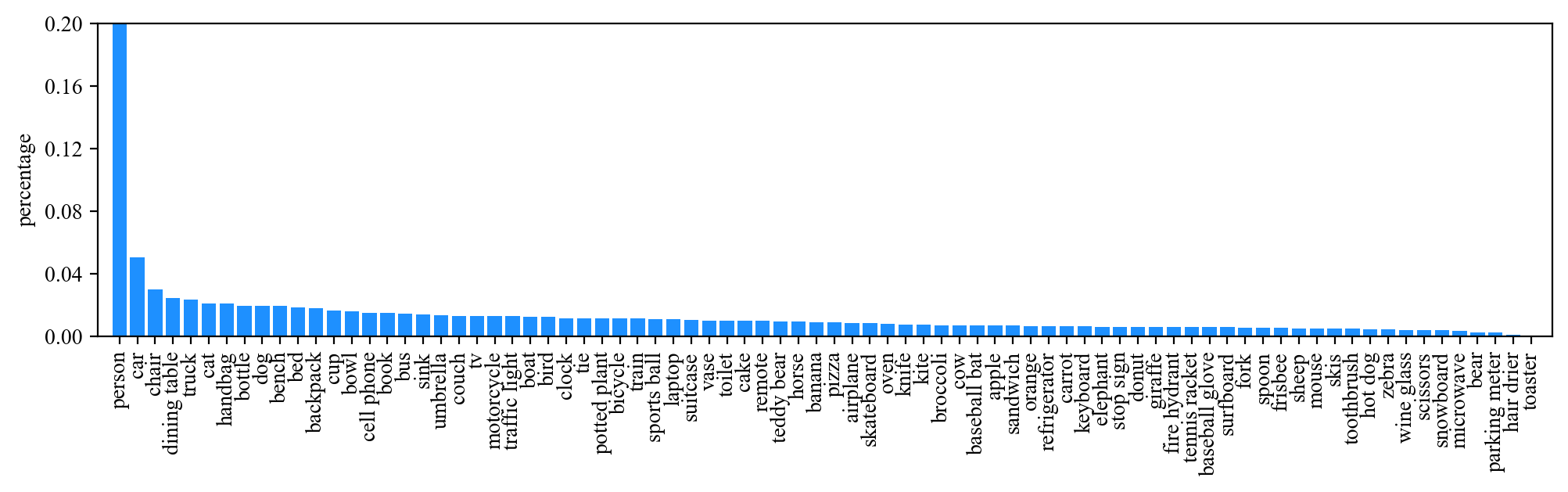}
\caption{Distribution of image numbers (\%) for each category in the Web-COCO dataset.}
\label{fig:distribution}
\end{figure}

\noindent\textbf{Web-COCO} contains 290,000 images, in which each image is assigned a set of labels according to the keywords. We randomly select 20,000 images for manual annotation to obtain more accurate and in-depth descriptions. 1) \textbf{Label noise.} Label noise is inevitable for the webly retrieved data. In the context of multi-label images, it can be divided into the following scenarios. First, an image may contain objects from categories that are not covered by the search keywords, which leads to false-negative labels. Second, a keyword category may be assigned to an image even though the corresponding object is absent, which leads to false-positive labels. Some examples that illustrate these two scenarios are presented in Figure~\ref{fig:samples}. To evaluate the reliability of the keyword-derived labels, we use the manual annotations of these 20,000 sampled images as ground-truth labels and regard the keyword-derived labels as predictions. We then compute the per-category precision and recall of the keyword labels, as shown in Figure~\ref{fig:pr}. As shown, the average precision and recall are 46.1\% and 64.6\%, respectively, suggesting that the label noise is quite severe. 2) \textbf{Semantic scattering.} Multi-label images contain multiple semantic objects scattered over the whole image. Thus, the model needs to locate the corresponding semantic regions to recover missing labels and examine the whole image to correct false-positive labels. 3) \textbf{Class imbalance.} Class imbalance naturally exists in real-world images, and it is more severe for webly retrieved multi-label images. As shown in Figure~\ref{fig:distribution}, the most frequent category ``person'' occupies about 15\% labels while the 20 least frequent categories occupy merely around 5\%. To evaluate the WS-MLR task, we use Web-COCO as the training set and the validation set of Microsoft COCO \cite{lin2014microsoft}, which contains 40,504 images with full manual annotations, as the validation set.

\begin{figure}
\centering
\includegraphics[width=0.8\textwidth]{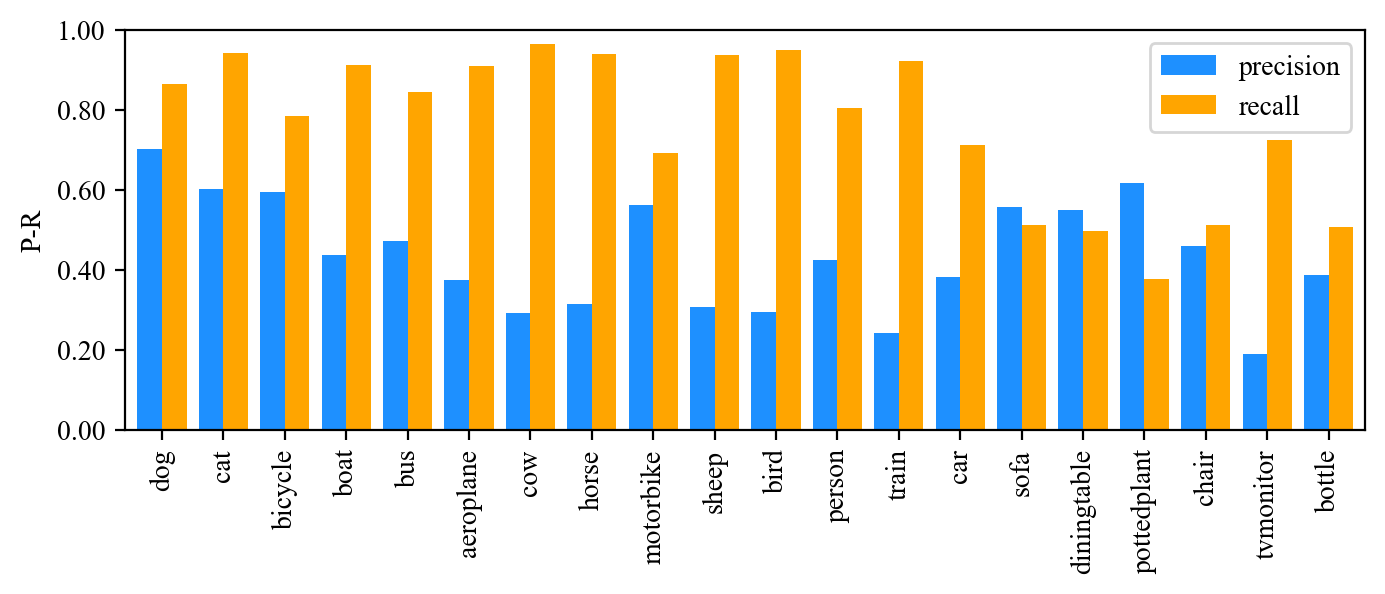}
\caption{Precision and recall of pseudo labels in the Web-Pascal dataset.}
\label{fig:pr-voc}
\end{figure}

\begin{figure}
\centering
\includegraphics[width=0.8\textwidth]{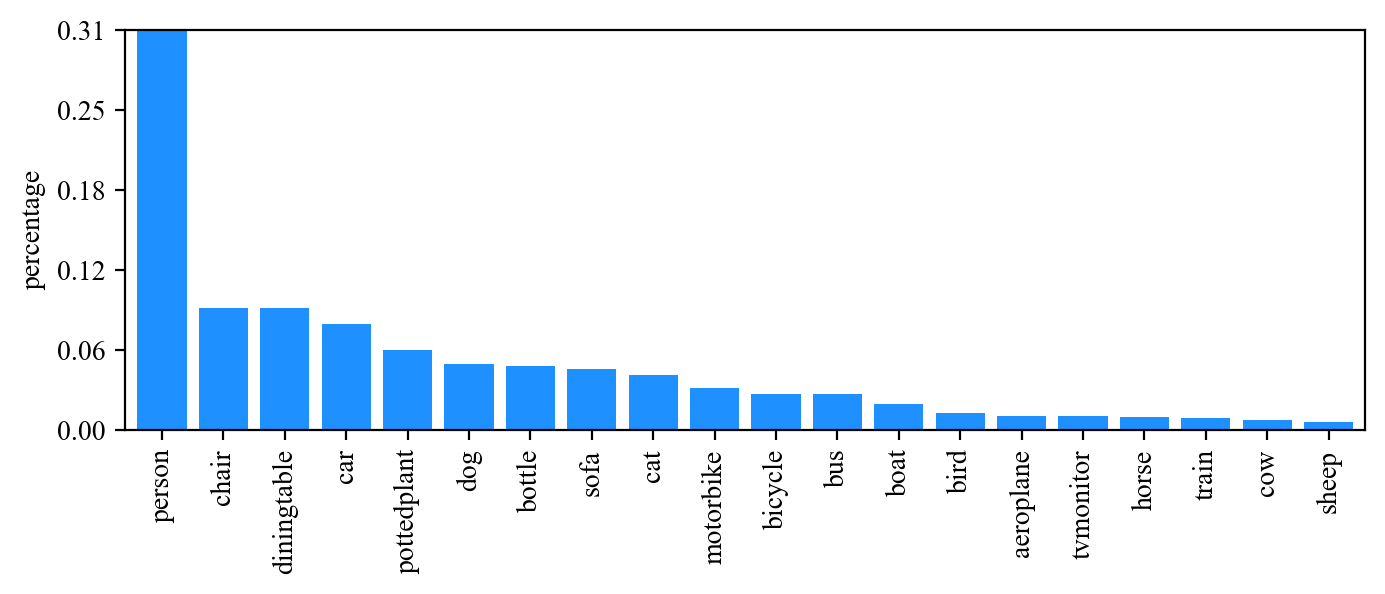}
\caption{Distribution of image numbers (\%) for each category in the Web-Pascal dataset.}
\label{fig:distribution-voc}
\end{figure}

\noindent\textbf{Web-Pascal} includes 236,043 images, and we use the keywords in the 20 categories of the Pascal VOC dataset \cite{everingham2010pascal} as its labels. As shown in Figures \ref{fig:pr-voc} and \ref{fig:distribution-voc}, this dataset also has the same challenges of label noise, semantic scattering, and class imbalance as Web-COCO. Similarly, Web-Pascal is used as the training set, and the testing set that contains 4,952 manually annotated images is used for evaluation.

It should be noted that Web-COCO and Web-Pascal are not fully manually labeled training sets. Instead, they are webly supervised multi-label training sets with keyword-derived noisy pseudo labels. For each image, we construct a $C$-dimensional label vector in the corresponding category space. The categories used as search keywords are treated as observed positive labels, while the remaining categories are treated as unobserved or assumed negative labels during training. Therefore, the training labels may contain both false-positive and false-negative entries, as illustrated in Figure~\ref{fig:samples}. For evaluation, we directly use the original manually annotated validation set of Microsoft COCO and Pascal VOC.

\begin{figure}
\centering
\includegraphics[width=0.99\textwidth]{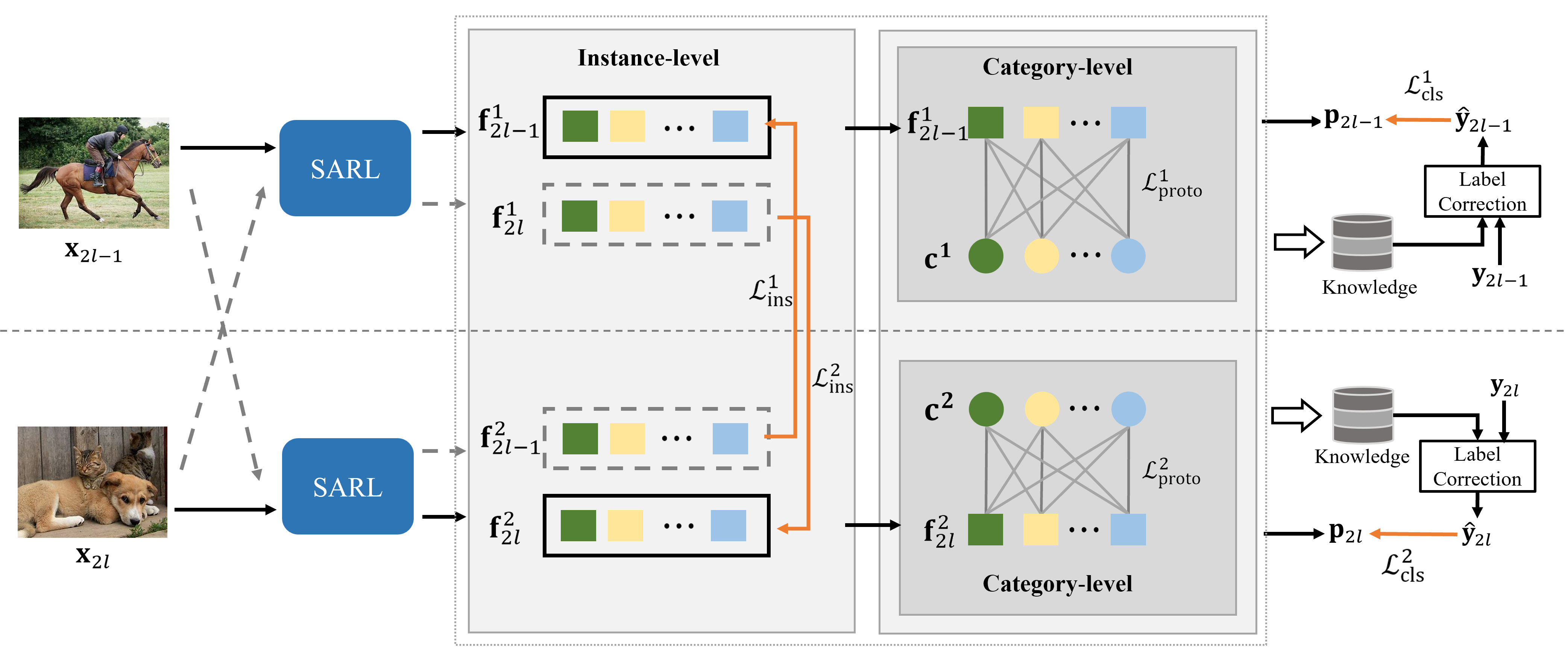}
\caption{Illustration of the DBMLCL framework. It contains two branches, and each of them consists of a feature extractor, a SARL module, and a classifier. The feature extractor extracts features from input images, and then the class-specific features are obtained by the SARL module. Finally, the prediction scores are output by the classifier. During training, we introduce instance-level and category-level contrastive losses to learn implicit knowledge representations, thus helping to correct label noise.}
\label{fig:framework}
\end{figure}

\section{Dual-Branch Multi-Label Contrastive Learning}
The DBMLCL framework employs two independent branches, each consisting of a backbone followed by a semantic-aware representation learning module. Each branch extracts category-specific instance-level feature vectors and maintains category-level prototypes. Two contrastive losses---instance-level and prototype-level---are introduced to learn compact representations and their similarities. These similarities are then used to detect and correct noisy labels. An overall illustration is presented in Figure~\ref{fig:framework}.

Given a training set with $N$ samples, we first divide them into two subsets, where each subset has $L=\frac{N}{2}$ non-overlapping samples, denoted as $\mathcal{D}^1=\{(\mathbf{x}_1, \mathbf{y}_1),(\mathbf{x}_3,\mathbf{y}_3),\ldots,(\mathbf{x}_{2L-1},\mathbf{y}_{2L-1})\}$ and $\mathcal{D}^2=\{(\mathbf{x}_2,\mathbf{y}_2),(\mathbf{x}_4,\mathbf{y}_4),\ldots,(\mathbf{x}_{2L},\mathbf{y}_{2L})\}$. $\mathbf{x}_i\in \mathbb{R}^{H\times W\times 3}$ is the input image of the $i$-th sample, where $H$ and $W$ are the height and width of the image. $\mathbf{y}_i\in \{0, 1\}^C$ is the pseudo label of the $i$-th sample, where $C$ is the category number. $\mathbf{y}_{i,c}$ is set to 1 if category $c$ is assumed to be present in the $i$-th image, and set to 0 otherwise.

Given an input image $\mathbf{x}_i$, the DBMLCL first uses two network branches $\mathcal{B}^1, \mathcal{B}^2$, where each branch $\mathcal{B}^j$ is used to extract instance-level feature vectors for all categories, denoted as 
\begin{equation}
    \mathbf{v}^j_i = [\mathbf{v}^j_{i,1},\mathbf{v}^j_{i,2},\ldots,\mathbf{v}^j_{i,C}] = \mathcal{B}^j(\mathbf{x}_i), j=1,2 
    \label{eq:feature}
\end{equation}
Here, we follow previous work \cite{chen2019learning} to implement $\mathcal{B}^j$ by a standard backbone (e.g., ResNet) followed by a semantic-aware representation learning module. The two branches $\mathcal{B}^1, \mathcal{B}^2$ share the same network architecture and do not share parameters. Based on the learned feature vectors, we also follow previous work to adopt a gated graph neural network and a fully connected layer to predict the confidence score for each category. Then, we apply the sigmoid function to compute the probability, denoted as 
\begin{equation}
    \mathbf{p}^j_i = \sigma(\phi_{cls}{(\mathbf{v}^j_i))}, j=1,2 
    \label{eq:prob}
\end{equation}

In the following, we introduce the instance and category contrastive losses, which promote learning category-specific instance-level and category-level feature representations together with their similarities. Then, we introduce the noisy label correction process.

\subsection{Instance Contrastive Loss}
The instance contrastive loss enforces that, for each positive category, the feature vectors extracted from the two branches for the same image are similar, while they are dissimilar to negative samples drawn from a circular queue.

Given a training image $\mathbf{x}_{i}$, we can first extract the feature vectors according to Equation~\ref{eq:feature}, obtaining feature vectors $\mathbf{v}^1_{i} = [\mathbf{v}^1_{i,1},\mathbf{v}^1_{i,2},\ldots,\mathbf{v}^1_{i,C}]$ and $\mathbf{v}^2_{i} = [\mathbf{v}^2_{i,1},\mathbf{v}^2_{i,2},\ldots,\mathbf{v}^2_{i,C}]$. For category $c$ existing in the image, the feature vectors $\mathbf{v}^1_{i,c}$ and $\mathbf{v}^2_{i,c}$ are expected to be similar and thus constitute a positive pair $(\mathbf{v}^1_{i,c}, \mathbf{v}^2_{i,c})$. For negative pairs, it is unreasonable to directly use feature vectors belonging to different categories from the same image because they may contain overlapping visual information. To address this issue, we further introduce a circular queue $\mathbf{Q}^1=[\mathbf{q}^1_1,\mathbf{q}^1_2,\ldots,\mathbf{q}^1_R]$, which is randomly initialized. After each iteration, we enqueue it with the feature vectors whose corresponding categories do not exist in the image. In this way, we can obtain negative pairs $(\mathbf{v}^1_{i,c},\mathbf{q}^1_{1,c}),(\mathbf{v}^1_{i,c},\mathbf{q}^1_{2,c}),\ldots,(\mathbf{v}^1_{i,c},\mathbf{q}^1_{R,c})$ for category $c$. Let $\mathcal{K}_i = \{c \mid y_{i,c}=1\}$ denote the set of positive categories for image $i$, and $|\mathcal{K}_i|$ be its cardinality. Then, the instance contrastive learning for branch $\mathcal{B}^1$ is defined as 
\begin{equation}
\begin{aligned}
    \mathcal{L}^1_{\text{inst}|i} &= -\frac{1}{|\mathcal{K}_i|}\sum_{c \in \mathcal{K}_i} \log \frac{\exp(\mathbf{v}^1_{i,c}\cdot \mathbf{v}^2_{i,c}/\tau)}{\exp(\mathbf{v}^1_{i,c}\cdot \mathbf{v}^2_{i,c}/\tau)+\sum_{r=1}^R \exp(\mathbf{v}^1_{i,c}\cdot \mathbf{q}^1_{r,c}/\tau)}
\end{aligned}
\label{eq:inst1}
\end{equation}
where $\tau$ is a temperature coefficient that is empirically set to 0.1 in our experiments.

Similarly, the instance contrastive learning for branch $\mathcal{B}^2$ is defined as 
\begin{equation}
\begin{aligned}
    \mathcal{L}^2_{\text{inst}|i'} &= -\frac{1}{|\mathcal{K}_{i'}|}\sum_{c \in \mathcal{K}_{i'}} \log \frac{\exp(\mathbf{v}^2_{i',c}\cdot \mathbf{v}^1_{i',c}/\tau)}{\exp(\mathbf{v}^2_{i',c}\cdot \mathbf{v}^1_{i',c}/\tau)+\sum_{r=1}^R \exp(\mathbf{v}^2_{i',c}\cdot \mathbf{q}^2_{r,c}/\tau)}
\end{aligned}
\label{eq:inst2}
\end{equation}
It is worth noting that the gradients of $\mathcal{L}^1_{\text{inst}|i}$ and $\mathcal{L}^2_{\text{inst}|i^{'}}$ are used to update the parameters of $\mathcal{B}^1$ and $\mathcal{B}^2$, respectively. In this way, the two network branches are updated based on different data, promoting them to learn diverse representations and thus facilitating contrastive learning.

\subsection{Prototype Contrastive Loss}
The prototype contrastive loss encourages each instance-level feature vector to be close to its corresponding category-level prototype and far from prototypes of other categories, thereby enhancing category-level discrimination.

We first learn the prototype representation $\mathbf{c}^j_c$ for each branch $j$ and category $c$. Concretely, each $\mathbf{c}^j_c$ is first initialized by a zero vector and then updated by the feature vectors of the corresponding category from the same branch as:
\begin{equation}
    \mathbf{c}^j_c \leftarrow m\mathbf{c}^j_c+(1-m)\mathbf{v}^j_{i,c}, \forall c \in \mathcal{K}_i, j=1,2
    \label{eq:momentum}
\end{equation}
where $m$ is a momentum factor, set to 0.999.

For each branch $j$, we constrain its feature vector $\mathbf{v}^j_{i,c}$ to be close to its own prototype $\mathbf{c}^j_c$ of the same category and far away from those of different categories. Thus, the prototype contrastive loss for branch $\mathcal{B}^1$ is defined as
\begin{equation}
\begin{aligned}
    \mathcal{L}^1_{\text{proto}|i} &= -\frac{1}{|\mathcal{K}_i|}\sum_{c \in \mathcal{K}_i} \log \frac{\exp(\mathbf{v}^1_{i,c}\cdot \mathbf{c}^1_c/\tau)}{\sum_{o=1}^C \exp(\mathbf{v}^1_{i,c} \cdot \mathbf{c}^1_o/\tau)}
\end{aligned}
\label{eq:proto1}
\end{equation}
Similarly, the prototype contrastive learning for $\mathcal{B}^2$ is defined as
\begin{equation}
\begin{aligned}
    \mathcal{L}^2_{\text{proto}|i'} &= -\frac{1}{|\mathcal{K}_{i'}|}\sum_{c \in \mathcal{K}_{i'}} \log \frac{\exp(\mathbf{v}^2_{i',c}\cdot \mathbf{c}^2_{c}/\tau)}{\sum_{o=1}^C \exp(\mathbf{v}^2_{i',c} \cdot \mathbf{c}^2_o/\tau)}
\end{aligned}
\label{eq:proto2}
\end{equation}

\subsection{Label Correction}
After warm-up training, when the model has learned reasonably discriminative representations, we use prediction scores to detect label noise and then generate new labels based on the similarity between semantic features and prototypes. Given image $\mathbf{x}_i$, we compute class probabilities $\mathbf{p}^1_i,\mathbf{p}^2_i$ by Equation~\ref{eq:prob}, and calculate the average probabilities $\mathbf{\overline{p}}_i = (\mathbf{p}^1_i + \mathbf{p}^2_i) / 2$. A label $\mathbf{y}_{i,k}$ is considered noisy if (i) $\mathbf{\overline{p}}_{i,k} > \theta_1$ while $\mathbf{y}_{i,k}=0$, or (ii) $\mathbf{\overline{p}}_{i,k} < \theta_2$ while $\mathbf{y}_{i,k}=1$. We encode these noisy labels with a binary mask $M_{noisy}$ as:
\begin{equation}
M_{noisy|i,k}=\left\{
    \begin{array}{ccc}
        1 & \mathbf{\overline{p}}_{i,k} > \theta_1\ \wedge\ \mathbf{y}_{i,k}=0 \\
        1 & \mathbf{\overline{p}}_{i,k} < \theta_2\ \wedge\ \mathbf{y}_{i,k}=1\\
        0 & \text{otherwise}
    \end{array}
\right.
\label{eq:noisy}
\end{equation}
For noisy labels, we correct them by embedding the knowledge of contrastive learning. Specifically, we generate new labels based on the cosine similarity between feature vectors and prototypes:

\begin{equation}
    \hat{\mathbf{s}}^j_{i,k}=\langle\mathbf{v}^j_{i,k},\mathbf{c}^j_k\rangle,j=1,2
\label{eq:similarity}
\end{equation}

The average similarity over the two branches is computed as:
\begin{equation}
    \mathbf{\overline{s}}_{i,k} = (\hat{\mathbf{s}}^1_{i,k} + \hat{\mathbf{s}}^2_{i,k}) / 2
\label{eq:avg_similarity}
\end{equation}

\begin{equation}
    \hat{\mathbf{y}}_{i,k}=\left\{
    \begin{array}{ccc}
        1 & \mathbf{\overline{s}}_{i,k} > \alpha_1 \wedge M_{noisy|i,k}=1\\
        0 & \mathbf{\overline{s}}_{i,k} < \alpha_2 \wedge M_{noisy|i,k}=1\\
        \mathbf{y}_{i,k} & \text{otherwise}
    \end{array}
\right.
\label{eq:y_proto}
\end{equation}
where $\langle x,y\rangle$ denotes the cosine similarity between $x$ and $y$, and $\mathbf{\overline{s}}_{i,k}$ is the average cosine similarity defined in Eq.~(\ref{eq:avg_similarity}). $\alpha_1,\alpha_2$ are the positive threshold and negative threshold, respectively. If the average similarity $\mathbf{\overline{s}}_{i,k}$ is greater than $\alpha_1$, the category is considered present; if the average similarity $\mathbf{\overline{s}}_{i,k}$ is less than $\alpha_2$, the category is considered absent; the pseudo label $\mathbf{y}_{i,k}$ is used in other cases. The pseudo-code for the label correction process is presented in Algorithm~\ref{alg:label_correction}.

To make the thresholds adaptive, we adopt category-specific thresholds and update them after every iteration through a momentum factor:
\begin{equation}
\begin{aligned}
    \alpha_{j,k} &\leftarrow  m\alpha_{j,k} + (1-m)\mathbf{\overline{s}}_{i,k},\\
    \theta_{j,k} &\leftarrow  m\theta_{j,k} + (1-m)\mathbf{\overline{p}}_{i,k},\\
    \forall i&\in \{i|\hat{\mathbf{y}}_{i,k} = 1\},j=1,2
\end{aligned}
    \label{eq:thresh}
\end{equation}
where we initialize $\theta_1$ and $\alpha_1$ to 1, $\theta_2$ and $\alpha_2$ to 0, and set $m$ to 0.999.

The momentum update is adopted to stabilize threshold estimation during training. Since the prediction scores and prototype similarities may fluctuate across mini-batches, directly using the current batch statistics or simple averaging can lead to unstable label correction. The momentum factor makes the thresholds evolve smoothly and reduces the influence of noisy mini-batch estimates. The initial values of 1 and 0 are conservative starting points rather than critical fixed thresholds: they prevent aggressive label correction at the beginning of training, and the thresholds are gradually adapted according to the model predictions, feature-prototype similarities, and data distribution. Therefore, the final thresholds are mainly determined by the training dynamics and dataset statistics rather than by fixed manually tuned initial values.

Once we have corrected the label noise and generated the new label $\hat{\mathbf{y}}_{i}$, we use the new label to compute losses and update the queue $\mathbf{Q}$ and prototype $\mathbf{c}$.

\begin{algorithm}
\caption{Label Correction}
\label{alg:label_correction}

\hspace*{0.02in} \textbf{Input:} Training images $\{\mathbf{x}_i\}_{i=1}^{N}$, keyword-derived pseudo labels $\{\mathbf{y}_i\}_{i=1}^{N}$, confidence thresholds $\theta_1,\theta_2$, similarity thresholds $\alpha_1,\alpha_2$, and class prototypes $\mathbf{c}^1,\mathbf{c}^2$\\
\hspace*{0.02in} \textbf{Output:} Rectified multi-label annotations $\{\hat{\mathbf{y}}_i\}_{i=1}^{N}$

\begin{algorithmic}[1]
\For{$i=1$ to $N$}
  \State Obtain feature vectors $\mathbf{v}_i^1,\mathbf{v}_i^2$ by Equation~\ref{eq:feature}
  \State Compute class probabilities $\mathbf{p}_i^1,\mathbf{p}_i^2$ by Equation~\ref{eq:prob}
  \State Compute average probabilities: $\mathbf{\overline{p}}_i \leftarrow (\mathbf{p}^1_i + \mathbf{p}^2_i) / 2$
  \State Initialize $\hat{\mathbf{y}}_{i}=\mathbf{y}_{i}$
  \For{$k=1$ to $C$}
    \If{($\mathbf{\overline{p}}_{i,k}>\theta_1$ and $\mathbf{y}_{i,k}==0$) or ($\mathbf{\overline{p}}_{i,k}<\theta_2$ and $\mathbf{y}_{i,k}==1$)}
      \State Mark $\mathbf{y}_{i,k}$ as a noisy label
      \State Compute average prototype similarity: 
      \State $\mathbf{\overline{s}}_{i,k}\leftarrow\left[\langle\mathbf{v}^1_{i,k},\mathbf{c}^1_k\rangle+\langle\mathbf{v}^2_{i,k},\mathbf{c}^2_k\rangle\right]/2$
      \State Rectify the noisy label based on $\mathbf{\overline{s}}_{i,k}$:
      \If$\mathbf{\overline{s}}_{i,k}>\alpha_1$
        \State $\hat{\mathbf{y}}_{i,k}\leftarrow 1$
      \ElsIf$\mathbf{\overline{s}}_{i,k}<\alpha_2$
        \State $\hat{\mathbf{y}}_{i,k}\leftarrow 0$
      \EndIf
    \EndIf
  \EndFor
\EndFor
\end{algorithmic}
\end{algorithm}

\subsection{Optimization}
We use binary cross-entropy loss as the classification objective function, which is defined as follows:
\begin{equation}
    \mathcal{L}_{\text{cls}|i}=-\frac{1}{C}\sum_{k=1}^C [y_{i,k}\log\mathbf{p}_{i,k}+(1-y_{i,k})\log(1-\mathbf{p}_{i,k})] \label{eq1:loss_cls}
\end{equation}
where $y_{i,k}$ denotes the current training label, which is the original pseudo label before label correction and the rectified label after label correction.
The final loss is the sum of the three losses, which is formulated as:
\begin{equation}
    \mathcal{L} = \mathcal{L}_{\text{cls}} + \lambda_1 \mathcal{L}_{\text{inst}} + \lambda_2 \mathcal{L}_{\text{proto}} \label{eq:loss_total}
\end{equation}
where $\lambda_1,\lambda_2$ are balance factors that are set to 0.01 and 0.05 respectively in the experiments.

During training, if the input image $\mathbf{x}_i \in \mathcal{D}^1$, we only compute losses and perform backpropagation for $\mathcal{B}^1$, and vice versa. We first train with pseudo labels for 5 epochs and then start label correction, using the generated labels for training. Applying label correction means that we trust the prediction results of the model more. Therefore, we set the weight decay parameter to 0 in our method to encourage the model to fit the corrected labels.

\section{Experiments}
In this section, we conduct extensive experiments based on the unified evaluation benchmark to evaluate the performance of the proposed DBMLCL framework and compare it with representative baselines. We also perform ablation studies to verify the effectiveness of each component to provide a better understanding of the DBMLCL framework.

\subsection{Experimental Settings}
\label{sec:exp_settings}

\subsubsection{Implementation Details} 

For a fair comparison, we adopt a unified training strategy for all algorithms. We use the Adam \cite{kingma2014adam} optimizer with an initial learning rate of $10^{-5}$, $\beta_1=0.9$, and $\beta_2=0.999$ to train for 20 epochs. The batch size is 8, and weight decay adopts the original settings in each paper. The learning rate is divided by 10 after 15 epochs of training. For data augmentation, the input image is resized to 512$\times$512 and cropped to a random size of \{512, 448, 384, 320, 256\}. Furthermore, the cropped patch is resized to 448$\times$448. Finally, the input image is randomly horizontally flipped and then normalized by ImageNet \cite{deng2009imagenet} pixel mean and variance. During inference, we average the prediction scores from the two branches of DBMLCL as the final prediction.

\begin{figure}
\centering
\includegraphics[width=0.99\textwidth]{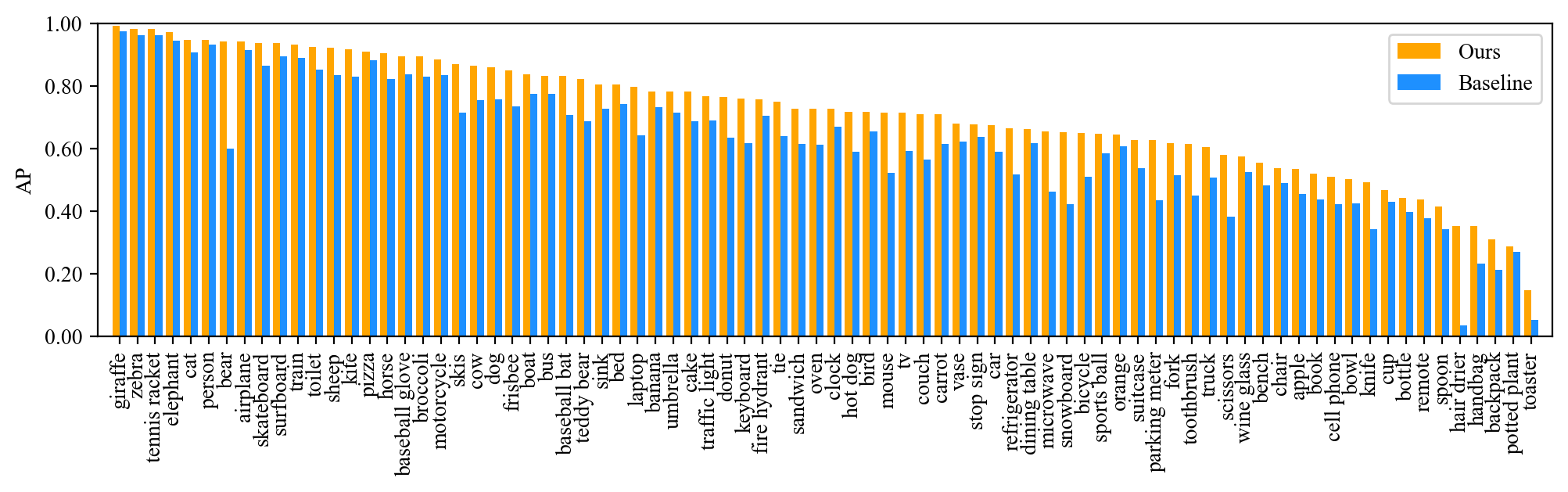}
\caption{The AP of each category of our proposed framework and the SSGRL baseline.}
\label{fig:ap_comparison}
\end{figure}

\subsubsection{Evaluation Metrics} 

We adopt the mean average precision (mAP) over all categories for evaluation. Moreover, we follow most previous MLR works \cite{chen2019learning,chen2019multi} to adopt the overall and per-class F1-scores (i.e., OF1 and CF1) for more comprehensive evaluation.

\subsection{Comparison with Representative Methods}
We evaluate the proposed DBMLCL against several representative baselines. These methods fall into two groups. The first group consists of traditional multi-label recognition algorithms, including SSGRL \cite{chen2019learning}, ML-GCN \cite{chen2019multi}, CSRA \cite{zhu2021residual}, ASL \cite{ridnik2021asymmetric}, P-GCN \cite{chen2021p-gcn}, and KGGR \cite{chen2022knowledge}. For these full-label methods, we use the complete $C$-dimensional keyword-derived pseudo label vector as supervision, where non-keyword categories are treated as negative labels. The second group addresses partial-label or limited-label scenarios; for P-GCN and KGGR, we follow their original protocols, treating non-keyword categories as unobserved when required. All methods use the same ResNet-101 \cite{he2016deep} backbone for fairness. We also include Classifier-guided CLIP Distillation (CCD) \cite{Kim_2025_CVPR}, a 2025 CLIP-based unsupervised multi-label classification method, as a modern vision-language baseline. All baselines are trained or adapted on the same web images and evaluated on the manually annotated validation/test sets to ensure fair comparison.

We follow the benchmark to present the performance comparisons and demonstrate the effectiveness of the DBMLCL framework over representative baselines.

\begin{table*}[htbp]
\centering
\caption{The mAP, OF1, and CF1 metrics of our DBMLCL framework and representative baselines on the Web-COCO dataset with different data proportions. The best results are highlighted in bold.}
\begin{tabular}{c|ccc|ccc|ccc}
\toprule
 \multirow{2}{*}{Methods} & \multicolumn{3}{c|}{20\%}  & \multicolumn{3}{c|}{40\%}  & \multicolumn{3}{c}{100\%} \\
\cmidrule(lr){2-4} \cmidrule(lr){5-7} \cmidrule(lr){8-10}
 & mAP & OF1 & CF1 & mAP & OF1 & CF1 & mAP & OF1 & CF1 \\
\hline
    SSGRL (2019)   & 62.0 & 58.9 & 55.5 & 63.1 & 59.1 & 55.8 & 64.6 & 61.1 & 57.3  \\
    ML-GCN (2019)  & 64.1 & 48.9 & 60.2 & 67.0 & 52.6 & 63.1 & 69.2 & 53.6 & 64.5  \\
    ASL (2021)    & 64.9 & 57.2 & 58.0 & 66.9 & 59.5 & 59.0 & 68.7 & 60.8 & 60.5  \\
    P-GCN (2021)  & 65.0 & 62.4 & 60.3 & 67.8 & 65.1 & 63.3 & 69.5 & 65.9 & 64.4  \\
    CSRA (2021)   & 64.9 & 62.5 & 61.4 & 67.5 & 64.1 & 63.3 & 70.3 & 66.2 & 65.5  \\
    KGGR (2022)   & 64.1 & 59.4 & 62.0 & 65.9 & 60.7 & 62.8 & 67.4 & 61.2 & 63.1  \\
    CCD (2025)    & 59.6 & 51.3 & 43.8 & 58.9 & 51.3 & 42.1 & 58.3 & 49.8 & 41.9 \\
    DBMLCL (Ours)  & \textbf{68.6} & \textbf{64.8} & \textbf{64.3} & \textbf{70.1} & \textbf{65.5} & \textbf{65.4} & \textbf{71.4} & \textbf{66.6} & \textbf{66.6}  \\
\bottomrule
\end{tabular}
\label{tab:web_coco}
\end{table*}

\begin{table*}[htbp]
\centering
\caption{The mAP, OF1, and CF1 metrics of our DBMLCL framework and representative baselines on the Web-Pascal dataset with different data proportions. The best results are highlighted in bold.}
\begin{tabular}{c|ccc|ccc|ccc}
\toprule
 \multirow{2}{*}{Methods} & \multicolumn{3}{c|}{20\%}  & \multicolumn{3}{c|}{40\%}  & \multicolumn{3}{c}{100\%} \\
\cmidrule(lr){2-4} \cmidrule(lr){5-7} \cmidrule(lr){8-10}
 & mAP & OF1 & CF1 & mAP & OF1 & CF1 & mAP & OF1 & CF1 \\
\hline  
    SSGRL (2019)   & 83.0 & 76.3 & 78.4 & 84.3 & 78.3 & 80.4 & 87.0 & 81.0 & 82.2  \\
    ML-GCN (2019) & 83.1 & 72.2 & 77.4 & 83.2 & 78.5 & 78.5 & 85.8 & 80.0 & 80.3  \\
    ASL (2021)    & 84.4 & 65.9 & 66.5 & 84.5 & 65.6 & 66.7 & 85.7 & 66.8 & 68.7  \\
    P-GCN (2021)  & 86.6 & \textbf{80.4} & 81.2 & 87.0 & 80.2 & 81.5 & 86.8 & 79.7 & 81.4  \\
    CSRA (2021)   & 82.7 & 76.3 & 77.1 & 83.3 & 76.8 & 77.9 & 85.4 & 78.9 & 79.9  \\
    KGGR (2022)   & 81.4 & 74.7 & 76.7 & 83.9 & 77.0 & 79.1 & 86.4 & 80.1 & 81.5  \\
    CCD (2025)    & \textbf87.9 & 52.7 & 58.4 & \textbf88.3 & 52.2 & 57.0 & \textbf87.7 & 52.8 & 60.6 \\
    DBMLCL (Ours) & 87.5 & 80.0 & \textbf{82.1} & 87.7 & \textbf{80.3} & \textbf{82.1} & 87.4 & \textbf{81.1} & \textbf{82.4}  \\
\bottomrule
\end{tabular}
\label{tab:web_pascal}
\end{table*}

\subsubsection{Comparison on Web-COCO} 
As shown in Table~\ref{tab:web_coco}, we present mAP, OF1, and CF1 metrics under the settings of 20\%, 40\%, and 100\% data proportions. Training images and labels are the same for all methods to ensure fairness. Existing best-performing conventional methods are CSRA and P-GCN, in which CSRA generates class-specific features through a spatial attention module and combines them with class-agnostic average pooling features, while P-GCN decomposes the visual representation of an image into a set of label-aware features and encodes the features into inter-dependent image-level prediction scores. The mAPs under the settings of 20\%-100\% data proportions are 64.9\%, 67.5\%, 70.3\% by CSRA, and 65.0\%, 67.8\%, 69.5\% by P-GCN. We also compare with CCD, a recent CLIP-based unsupervised MLR method. Its performance on Web-COCO is limited, which may be attributed to the mismatch between its unsupervised pseudo-label distillation strategy and the WS-MLR training data. CCD mainly depends on CLIP-derived image-text rankings to generate supervision, while Web-COCO contains diverse web images with off-topic retrieval results, missing co-occurring objects, and noisy keyword labels. In this setting, CLIP pseudo labels are difficult to refine reliably, and the informative keyword supervision provided by the web search process is not explicitly exploited. Different from these methods, our framework DBMLCL learns category-specific instance-level and category-level representations as well as the similarity metrics to correct label noise, leading to a notable performance improvement on all metrics. Specifically, it achieves the best performance at all data proportions on Web-COCO with mAP of 68.6\%, 70.1\%, and 71.4\%, outperforming the second-best methods by 3.6\%, 2.3\%, and 1.1\%, respectively. DBMLCL also achieves the best OF1 of 64.8\%, 65.5\%, 66.6\%, and the best CF1 of 64.3\%, 65.4\%, 66.6\%. These comparisons demonstrate the effectiveness of the proposed DBMLCL framework on Web-COCO.

\subsubsection{Comparison on Web-Pascal} 

Compared with Web-COCO, Web-Pascal is a relatively simple dataset with fewer categories. The mAP, OF1, and CF1 results on Web-Pascal are also shown in Table~\ref{tab:web_pascal}. CCD obtains the highest mAP on Web-Pascal, which is reasonable because the 20 Pascal VOC categories are common object concepts and are better aligned with CLIP's web-scale image-text pretraining. This allows CCD to rank positive categories more accurately. However, mAP mainly evaluates ranking quality, while OF1 and CF1 depend on the final multi-label decision quality. CCD still produces much lower OF1 and CF1 scores, suggesting that its CLIP-based pseudo labels are not sufficiently calibrated for precise multi-label prediction under the WS-MLR protocol. In contrast, DBMLCL achieves more balanced and robust performance. It obtains the best CF1 of 82.1\%, 82.1\%, and 82.4\%, and the best OF1 at the 40\% and 100\% data proportions. These comparisons show that explicit category-specific noise correction remains important even when strong vision-language priors are introduced.

\subsection{Computational Efficiency Analysis}

To further analyze the computational cost of different methods, we report the number of parameters, FLOPs per image, training throughput, and peak memory usage in Table~\ref{tab:efficiency}. The FLOPs are computed with an input image size of $448\times448$, and the training throughput and peak memory are measured on an Ascend 910B NPU under the same measurement protocol. We do not report inference latency because deployment-oriented inference optimization is highly dependent on hardware devices, inference frameworks, operator implementations, and batching strategies. In comparison, FLOPs per image provides a more objective indicator of the inference computational cost.

\begin{table*}[htbp]
\centering
\caption{Computational efficiency comparison of representative methods. FLOPs are computed with an input image size of $448\times448$, and training throughput and peak memory are measured on an Ascend 910B NPU.}
\begin{tabular}{c|cccc}
\toprule
Method & Params (M) & FLOPs/Image (G) & Train Throughput (img/s) & Peak Memory (GB) \\
\hline
SSGRL & 92.31 & 90.61 & 70.11 & 2.88 \\
ML-GCN & 44.93 & 62.67 & 77.50 & 0.72 \\
ASL & 42.66 & 62.67 & 82.58 & 0.61 \\
P-GCN & 46.87 & 62.67 & 81.58 & 0.96 \\
CSRA & 43.16 & 62.93 & 79.14 & 0.62 \\
KGGR & 68.74 & 74.74 & 82.81 & 1.69 \\
CCD & 42.66 & 62.73 & 46.20 & 5.77 \\
DBMLCL & 193.53 & 182.66 & 16.67 & 4.78 \\
\bottomrule
\end{tabular}
\label{tab:efficiency}
\end{table*}

Compared with the SSGRL baseline, DBMLCL has approximately twice the number of parameters and FLOPs, which is expected because it extends SSGRL into a dual-branch framework. Its training throughput is lower than those of the single-branch baselines because training DBMLCL requires two-branch forward computation, contrastive feature collection, prototype updating, and label correction. These additional costs are mainly introduced during training to obtain more reliable category-specific representations and corrected labels. Therefore, DBMLCL provides improved WS-MLR performance at the cost of increased training computation.

\subsection{Ablation Study}

\begin{table}[htbp]
\centering
\caption{mAP, OF1, and CF1 of the SSGRL baseline, Dual-Branch SSGRL baseline, Dual-Branch SSGRL baseline with the instance contrastive loss and prototype contrastive loss, and our DBMLCL. All experiments are conducted under the setting of 100\% data proportion on the Web-COCO dataset.}
\begin{tabular}{c|ccc}
\toprule
Methods & mAP & OF1 & CF1 \\
\hline
SSGRL & 64.6 & 61.1 & 57.3 \\
Dual-Branch SSGRL & 64.2 & 59.2 & 54.5 \\
Dual-Branch SSGRL w/ $\mathcal{L}_\text{inst}$ & 66.8 & 62.5 & 60.0 \\
Dual-Branch SSGRL w/ $\mathcal{L}_\text{proto}$ & 65.8 & 63.4 & 61.6 \\
Ours & 71.4 & 66.6 & 66.6 \\
\bottomrule
\end{tabular}
\label{tab:ablation}
\end{table}

\subsubsection{Effectiveness of Dual-Branch Multi-Label Contrastive Learning}
The proposed framework builds on the SSGRL and introduces a dual-branch multi-label contrastive learning algorithm to help correct the noisy labels. To verify its contribution, we emphasize the comparisons with SSGRL in Table~\ref{tab:ablation}. As shown, under the 100\% data setting, DBMLCL improves the mAP, OF1, and CF1 from 64.6\%, 61.1\%, and 57.3\% to 71.4\%, 66.6\%, and 66.6\%, with increases of 6.8\%, 5.5\%, and 9.3\%, respectively. We also present the AP of each category in Figure~\ref{fig:ap_comparison} for a more detailed comparison. On the other hand, the proposed framework uses two network branches. Thus, we also carry out another baseline that trains two-branch SSGRL networks under the same setting as the DBMLCL and aggregates both outputs for final predictions. As shown in Table~\ref{tab:ablation}, simply using two SSGRL branches does not improve the performance over the single-branch SSGRL baseline, while DBMLCL achieves substantially better results. This indicates that the gain comes from the proposed contrastive learning and label correction design rather than merely increasing the number of branches.

\begin{figure}
\centering
\includegraphics[width=0.99\textwidth]{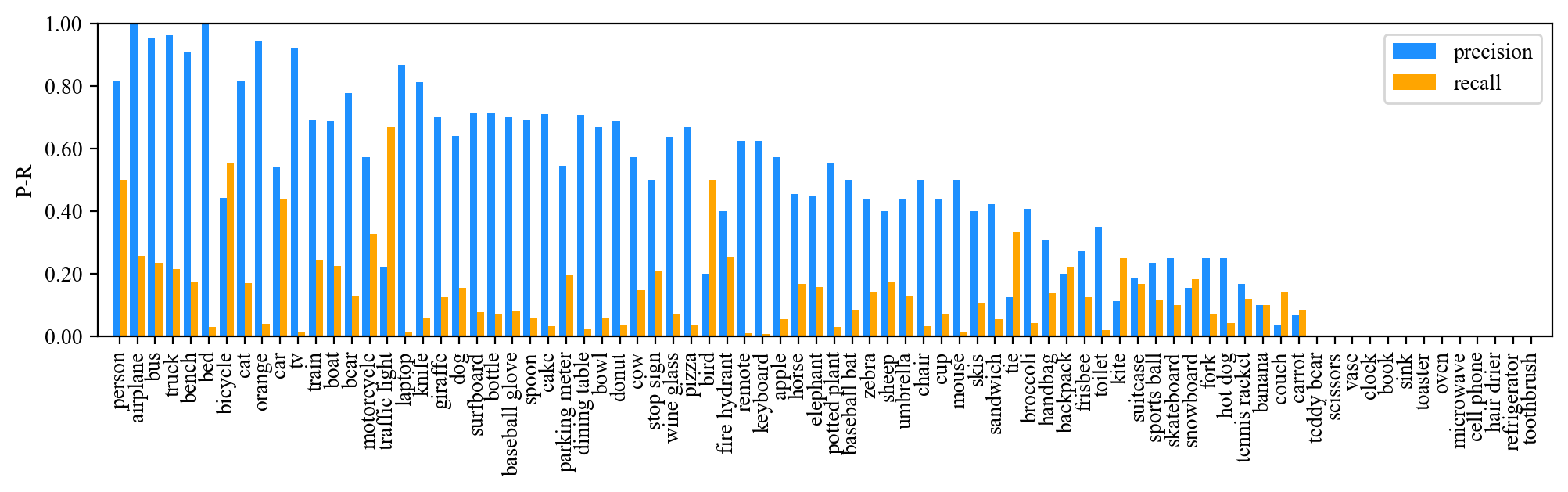}
\caption{Precision and recall of the retrieved labels produced by the label correction algorithm.}
\label{fig:retrieved_pr}
\end{figure}

\subsubsection{Analysis of Instance Contrastive Loss} 
Instance contrastive learning helps the two branches maintain consistency for the features of the same sample and widens the distance between the features of different samples, thus promoting the learning of more compact and discriminative features. Here, we verify its contribution by only adding it to the Dual-Branch SSGRL baseline. As shown in Table~\ref{tab:ablation}, it improves the mAP, OF1, and CF1 by 2.6\%, 3.3\%, and 5.5\%, respectively, under the 100\% data setting, which demonstrates its effectiveness. 

\subsubsection{Analysis of Prototype Contrastive Loss}
We also add prototype contrastive learning $\mathcal{L}_\text{proto}$ to the baseline Dual-Branch SSGRL to evaluate its contribution. Prototype contrastive learning helps features from the same category to be close to each other and features from different categories to be far away from each other, making the features more discriminative at the category level. Therefore, we can judge whether features belong to the same category based on their inter-class distance. As shown in Table~\ref{tab:ablation}, adding $\mathcal{L}_\text{proto}$ to Dual-Branch SSGRL improves the mAP, OF1, and CF1 by 1.6\%, 4.2\%, and 7.1\%, respectively, under the 100\% data setting. 

\subsubsection{Analysis of Label Correction}

To evaluate the efficacy of the label correction algorithm, we analyze the rectified labels produced by the label correction algorithm based on 20,000 annotated samples. The precision and recall metrics for the retrieved labels are depicted in Figure~\ref{fig:retrieved_pr}. Specifically, for the ``person'' category, we can observe a recall of approximately 50\% and a precision of 80\%. Similarly, the ``airplane'' category shows a recall exceeding 22\% and a precision nearing 100\%. This indicates that the algorithm effectively rectifies some noisy labels within web datasets. The relatively low recall for the ``person'' category is mainly caused by its high intra-class diversity and frequent co-occurrence with other categories. In web images, persons often appear as accompanying or background objects rather than search-keyword objects, and they may vary greatly in pose, scale, occlusion, and visible body parts. Therefore, many true ``person'' labels are missing from the keyword-derived labels but are difficult to retrieve confidently. Our label correction strategy is intentionally conservative: it retrieves a missing label only when both the prediction confidence and feature-prototype similarity provide reliable evidence. This leads to high precision for retrieved ``person'' labels, but some ambiguous or low-confidence person instances are not recovered, resulting in lower recall.

\begin{figure}
    \centering
    \includegraphics[width=0.7\textwidth]{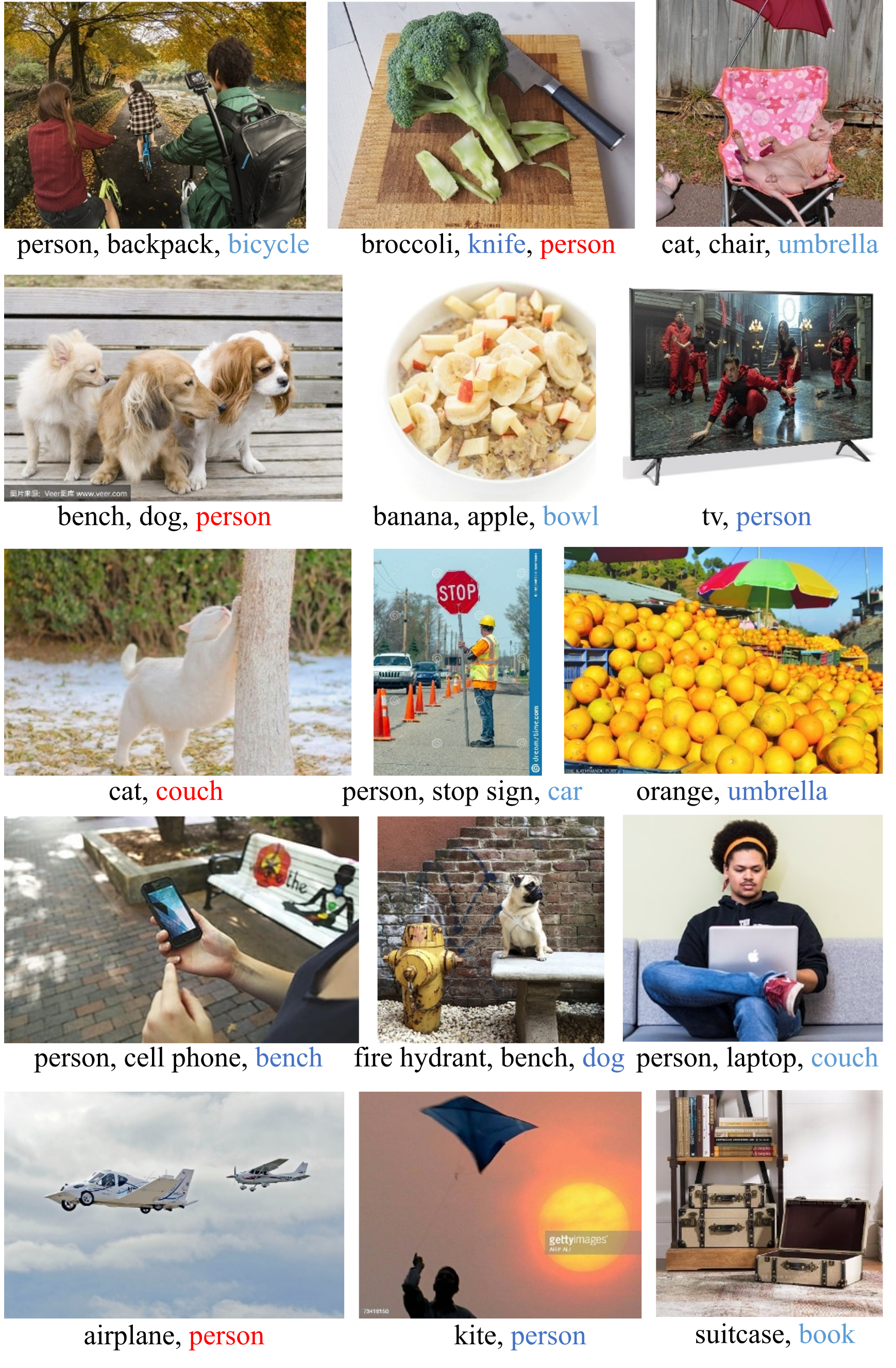}
    \caption{Some examples of the rectified labels produced by the label correction algorithm, with retrieved labels in blue and discarded labels in red.}
    \label{fig:rectified_examples}
    \end{figure}

We further elucidate the label correction algorithm with visual illustrations of the rectified labels, as presented in Figure~\ref{fig:rectified_examples}, to facilitate intuitive verification. For instance, in the first image, the initial pseudo labels are ``person'' and ``backpack''. Our algorithm effectively recognizes and retrieves the omitted label ``bicycle''. Similarly, in the second image, while the initial pseudo labels are ``broccoli'' and ``person'', our algorithm corrects the false-positive label ``person'' and retrieves the overlooked label ``knife''.

We also observe several typical failure cases. First, labels of small, heavily occluded, or partially visible objects may still be missed because their category-specific features are not sufficiently reliable for confident retrieval. Second, semantically or visually similar categories, such as ``cup'' and ``bottle'', ``chair'' and ``couch'', or ``knife'' and ``fork'', may lead to ambiguous feature-prototype similarities and cause incorrect rectification. Third, web images sometimes contain complex backgrounds, product-style images, posters, or off-topic retrieval results, where the search keywords are only weakly related to the actual visual content. These cases remain challenging for label correction and indicate that more fine-grained visual grounding or stronger vision-language priors may further improve WS-MLR.

\begin{table}[htbp]
    \centering
    \caption{Performance comparison under different training data partitions for each branch.}
    \begin{tabular}{c|ccc}
    \toprule
    Data per branch & mAP & OF1 & CF1 \\
    \hline
    50\% & 70.1 & 65.5 & 65.4 \\
    75\% &  70.0 & 65.7 & 65.1 \\
    100\% & 70.0 & 65.9 & 65.2  \\
    \bottomrule
    \end{tabular}
    \label{tab:data_partition}
    \end{table}

\subsubsection{Analysis of Training Data Partition for Each Branch}
The two branches of DBMLCL use different training data to construct instance contrastive learning, which can introduce additional clean supervision signals to allow the two branches to learn from each other, thereby enhancing model performance. Here we discuss the data partition strategy and test the performance when each branch has 50\%, 75\%, and 100\% training data. As shown in Table~\ref{tab:data_partition}, there is little difference in the performance of the three cases. To reduce training consumption, we set 50\% training data in each branch of DBMLCL.

It should be noted that the 50\%/50\% setting does not discard half of the training data. The two branches are trained on non-overlapping subsets whose union covers the whole training set. When the data proportion for each branch increases to 75\% or 100\%, the additional samples mainly introduce overlap between the two branches rather than new unique web images. Therefore, the marginal gain is limited. Moreover, in WS-MLR, the main bottleneck comes from noisy keyword-derived labels and missing labels, so repeatedly exposing both branches to more overlapping noisy samples does not necessarily improve the learned representations. The similar performance across different partitions suggests that 50\%/50\% provides a good trade-off between branch diversity, data coverage, and training efficiency.

\section{Conclusion}
In this work, we study webly supervised multi-label recognition (WS-MLR), a practical yet challenging setting where large-scale web images are used as training data and keyword-derived labels contain both false-positive and false-negative noise. We construct a unified evaluation benchmark by building two webly retrieved datasets, Web-COCO and Web-Pascal, and by re-implementing representative baselines under the same training and testing protocol. This benchmark provides a reproducible platform for evaluating how different MLR methods behave when moving from clean manually annotated data to noisy web supervision.

To address the label noise in WS-MLR, we propose a Dual-Branch Multi-Label Contrastive Learning framework that adapts contrastive learning and label correction to category-specific multi-label recognition. By learning category-specific instance-level representations, category-level prototypes, and their similarities, DBMLCL corrects noisy web labels at the category level rather than relying only on holistic image representations. Extensive experiments on Web-COCO and Web-Pascal show that DBMLCL achieves strong and balanced performance across mAP, OF1, and CF1 compared with conventional MLR methods, weakly supervised MLR methods, and recent CLIP-based multi-label recognition. These results demonstrate the importance of explicitly modeling category-specific label noise in WS-MLR.

Nevertheless, this study still has several limitations. First, the dual-branch training strategy introduces additional training cost compared with single-branch methods. Second, although Web-COCO and Web-Pascal provide useful WS-MLR benchmarks, webly retrieved data may still contain search-engine bias, long-tailed category coverage, off-topic images, and language-dependent retrieval noise. Third, our framework currently uses vision-language models mainly for comparison rather than deeply integrating their semantic priors into noise correction. In future work, we will further investigate the computational efficiency of DBMLCL, explore more efficient training strategies, and incorporate stronger vision-language priors for noisy web multi-label data.

\printcredits

\section*{Declaration of competing interest}
The authors declare that they have no known competing financial interests or personal relationships that could have appeared to influence the work reported in this paper.

\section*{Data availability}
The training datasets Web-COCO and Web-Pascal used in this study are available at Baidu Cloud: \url{https://pan.baidu.com/s/1Ipue3jpsFfqcUOf8JZJBTw?pwd=hjt3}.

\bibliographystyle{cas-model2-names}

\bibliography{cas-refs}



\end{document}